\def\year{2019}\relax
\documentclass[letterpaper]{article} 
\usepackage{aaai19}  
\usepackage{times}  
\usepackage{helvet}  
\usepackage{courier}  
\usepackage{url}  
\usepackage{graphicx}  

\usepackage{amsfonts}       
\usepackage{amsmath}
\usepackage{amssymb}
\usepackage{bbm}
\usepackage{bm}
\usepackage{algorithm}
\usepackage{algorithmic}
\usepackage{adjustbox}

\usepackage{subcaption}

\newcommand{\Wmat}{{\bf W}}

\newcommand{\cv}{{\boldsymbol c}}

\newcommand{\gv}{{\boldsymbol g}}
\newcommand{\hv}{{\boldsymbol h}}

\newcommand{\sv}{{\boldsymbol s}}

\newcommand{\vv}{{\boldsymbol v}}

\newcommand{\xv}{{\boldsymbol x}}

\newcommand{\R}{\mathbb{R}}

\newcommand{\Lcal}{\mathcal{L}}

\begin{document}
%
\title{Hierarchically Structured Reinforcement Learning for Topically Coherent Visual Story Generation}
\author{Qiuyuan Huang$^{1*}$, Zhe Gan$^{1}$\thanks{Equal Contribution.}, Asli Celikyilmaz$^1$, Dapeng Wu$^2$, Jianfeng Wang$^1$, Xiaodong He$^3$\\
$^1$Microsoft Research, Redmond, WA, $^2$University of Florida, $^3$JD AI Research\\
\{qihua, zhe.gan, aslicel, jianfw\}@microsoft.com; dpwu@ieee.org; xiaodong.he@jd.com
}
\maketitle
\begin{abstract}
We propose a hierarchically structured reinforcement learning approach to address the challenges of planning for generating coherent multi-sentence stories for the visual storytelling task. Within our framework, the task of generating a story given a sequence of images is divided across a two-level hierarchical decoder. The high-level decoder constructs a plan by generating a semantic concept (\emph{i.e.}, topic) for each image in sequence. The low-level decoder generates a sentence for each image using a semantic compositional network, which effectively grounds the sentence generation conditioned on the topic. The two decoders are \textit{jointly} trained end-to-end using reinforcement learning. We evaluate our model on the visual storytelling (VIST) dataset. Empirical results from both automatic and human evaluations demonstrate that the proposed hierarchically structured reinforced training achieves significantly better performance compared to a strong flat deep reinforcement learning baseline.
\end{abstract}

\section{Introduction} \label{sec:Introduction}
Visual storytelling is the task of generating a sequence of coherent sentences (\emph{i.e.}, a story) for an ordered image stream~\cite{park2015expressing,huang2016visual,liu2017let}. Inspired by the successful use of recurrent neural network (RNN) based encoder-decoder models employed in machine translation tasks~\cite{cho2014learning,sutskever2014sequence}, variants of encoder-decoder models have shown promising results on the task of story generation~\cite{huang2016visual}.

\begin{figure*}[t!]
	\centering
	\includegraphics[width=0.89\textwidth]{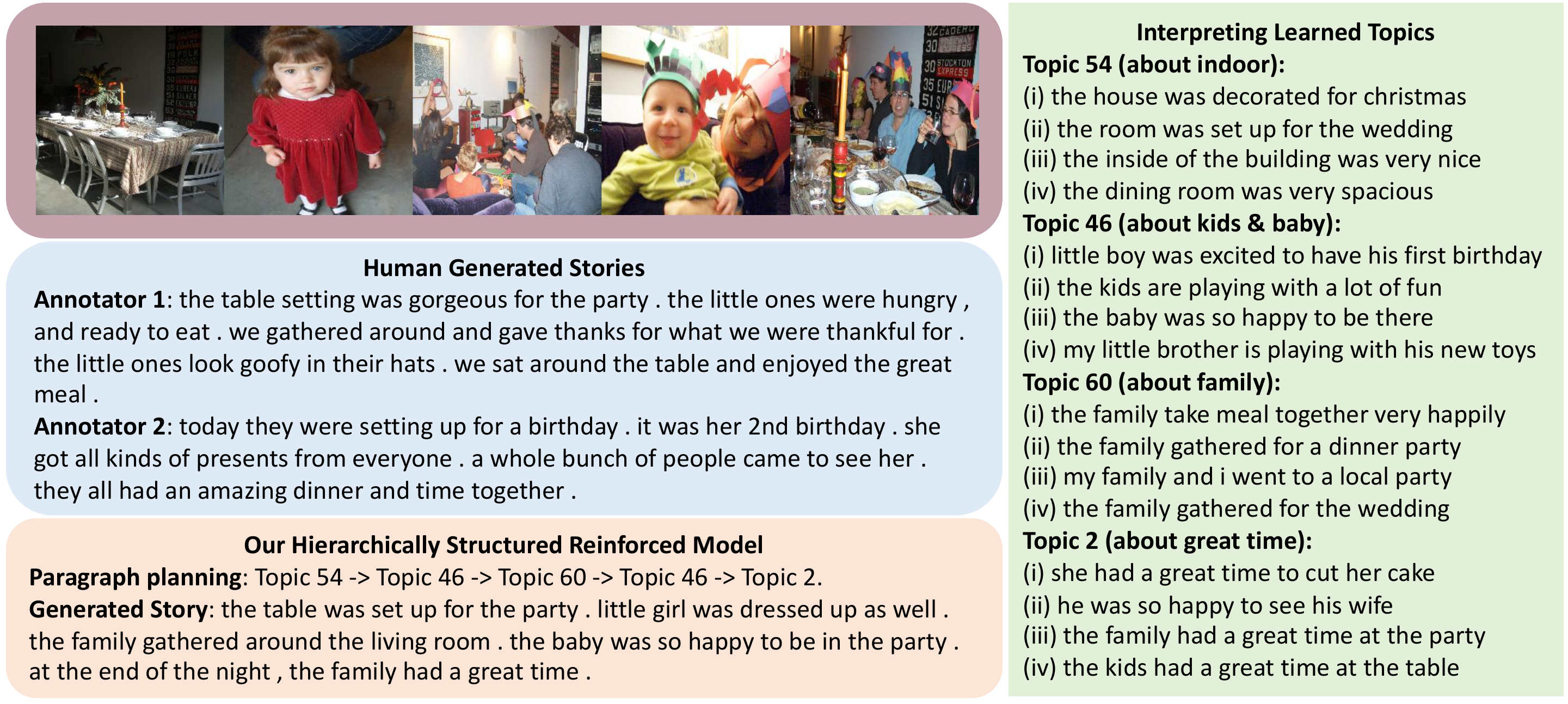}
	\caption{Example of hierarchically structured reinforcement learning for visual storytelling. Our model generates coherent stories by paragraph planning, \emph{i.e.}, predicting a sequence of topics. In order to visualize learned topics, we present sentences generated from the corresponding topics in the test set. We manually assigned the topic names in this example for visual clarity.
		\small }
	\vspace{-1mm}
	\label{fig:framework}
\end{figure*}

The fundamental challenge, however, is that the strong performance of neural encoder-decoder models does not generalize well for visual storytelling.  
The task requires a full understanding of the content of each image as well as the relation among different images.
The motivation behind our approach is to build a context-aware text-synthesis model that can efficiently encode the sequence of images and generate a topically coherent multi-sentence paragraph (see Fig.~\ref{fig:framework}). We design a two-level hierarchical structure, where a high-level decoder 
constructs a plan by generating a topic for each image in sequence, and the low-level decoder generates a sentence conditioned on that topic. 

Although the maximum likelihood estimation (MLE) is commonly used as the training loss for encoder-decoder RNNs, it may not be an appropriate surrogate for coherent long span generation task such as story generation. By only maximizing the ground truth probability, MLE can easily fail to exploit the wealth of information offered by the task specific losses. Most recent work in image captioning use reinforcement learning (RL) by providing sentence-level evaluation metrics for RNN training using global reward signals~\cite{Rennie2016Self,ren2017deep,liu2016improved} (e.g, BLEU score). Motivated by the success of these work, we use RL to train our hierarchically structured model.

\begin{figure}[t]
	\centering
	\includegraphics[width=1.0\linewidth]{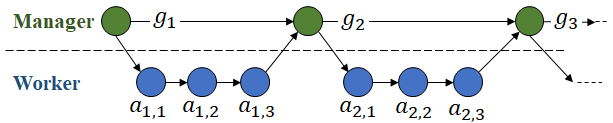}
	\caption{\small Overview of the Hierarchically Structured Reinforcement Learning, which consists of a Manager, a high-level decoder that generates topic (subgoal) sequences {$g_1$,$g_2$,..}, and a Worker, a low-level decoder which generates word sequences $a_{l,1}$,..$a_{l,T}$ conditioned on the selected topic $g_l$.}
	\label{fig:overview_hrl_agent}	
\end{figure}

More specifically, we propose the hierarchically structured reinforcement learning (HSRL) model to realize a two-level generation mechanism (schematic overview is shown in Fig.~\ref{fig:overview_hrl_agent}). Our model consists of a high-level decoder (\emph{i.e.}, Manager) and a low-level decoder (\emph{i.e.}, Worker). The Manager aims to produce a sequence of semantically coherent topics for an image stream so that the overall theme is distributed among each sentence in the generated story. The topics assigned by the Manager are then supplied to the Worker for performing the task of sentence generation, given the surrounding image and textual context. Our Manager is a long short-term memory (LSTM) network~\cite{hochreiter1997long}, while the Worker is a semantic compositional network (SCN)~\cite{SCN_CVPR2017}, which effectively incorporates the topical information into the sentence generation process. The Manager and Worker are trained end-to-end jointly using a mixed MLE and self-critical reinforcement learning loss~\cite{Rennie2016Self} to generate focused and coherent stories. 

Empirical results on the VIST dataset from both automatic and human evaluations demonstrate that the two-level decoder structure helps generate stories with improved narrative quality due to the control of overall thread on the generation of each sentence. 
Our benchmark analysis show that the hierarchically structured RL model significantly outperforms a strong flat deep reinforcement learning baseline, showing the effectiveness of paragraph planning and reinforcing the storytelling task at different levels.

\section{Related work}\label{sec:RelatedWork}
Early work on generating descriptions for images have shaped the field of image/video captioning. A typical model extracts a visual feature vector via a CNN, and then sends it to a language model for generation of a single sentence caption. Most notable work includes~\cite{vinyals2015show,xu2015show,fang2015captions,donahue2015long,karpathy2015deep} for image captioning, and~\cite{venugopalan2015sequence,pan2015hierarchical,yu2016video,pan2016joint,pu2018adaptive} for video captioning. 

More recently, the field has emerged into generation of long form text with the introduction of image paragraph generation~\cite{krause2016hierarchical} and dense video captioning~\cite{krishna2017dense} tasks. In this work, we focus on visual storytelling, investigating the generation of narrative paragraph for a photo stream. First initial models used sequence-to-sequence framework~\cite{huang2016visual,liu2017let}, while a joint embedding model was further developed in~\cite{liu2017let} to overcome the large visual variance issue in image streams. Later in~\cite{yu2017hierarchically}, the task of album summarization and visual storytelling are jointly considered. While we share the same motivation as the above previous work, all of them rely on MLE training leaving out the fundamental problems, \emph{e.g.}, exposure bias~\cite{bengio2015scheduled} or not optimizing for the desired objective, which we tackle in this paper.

Most recent work on training captioning and story generation has used sequence scores such as BLEU~\cite{ranzato2015sequence} or CIDERr~\cite{Rennie2016Self}  as a global reward to train a policy with the REINFORCE algorithm~\cite{williams1992simple}. These work mainly focus on single sentence generation using \emph{flat} RL approaches. In contrast, our work uses a hierarchically structured RL framework for capturing higher level semantics of the story generation task.

Our hierarchically structured model is related to the HRL work~\cite{wang2018video} for video captioning; however, they have not explored the discovery or the usage of interpretable subgoals. The main novelty of our work when compared to them is the usage of explicit topics as subgoal representations. This yields significant improvements over the baselines, as well as provides a clear semantic subgoal for the sentences to be generated. In addition, one of the other novelty of our work is the usage of the SCN as the Worker, rather than a flat LSTM. Further, we introduce new approaches for training the high- and low-level decoders together, rather than in an iterative manner.

Our work is also related to~\cite{wang2018show,wang2018no}, which uses adversarial training and inverse RL, respectively, for storytelling. However, neither of them has explored modeling of an explicit paragraph planning procedure. We introduced a ``plan-ahead'' strategy by using learned topics and proposing a hierarchically structured RL approach.

\section{Hierarchically Reinforced Generation} \label{sec:model_description}
\begin{figure*}[t]
	\centering
	\includegraphics[width=0.8\linewidth]{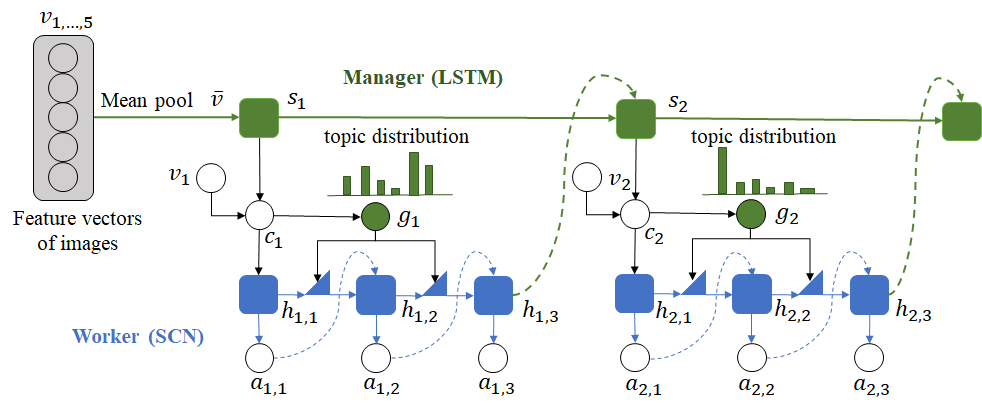}		
	\caption{\small Proposed Manager-Worker framework. For each sequence of images, the Manager LSTM (top) generates a topic distribution $g_{\ell}$, and the Worker SCN (Semantic Compositional Network) (bottom) generates sentences word by word $a_{\ell,t}$ conditioned on the topic distribution. Dash lines indicate copy operation where the output of one node is copied as input to the next node.}
	\label{fig:implementation_hrl_agent}
\end{figure*}

Recent work in image captioning ~\cite{Rennie2016Self,Pasunuru2017Reinforced}, machine translation~\cite{wu2016google}, and summarization~\cite{paulus2017deep,Celikyilmaz2018DCA} describe the benefits of fine-tuning neural generation systems with policy gradient methods on sentence-level scores (e.g. BLEU or CIDEr). 
While these approaches are able to learn a rough approximation of language when producing short sequences, they struggle on tasks involving long sequences (e.g., summaries, image paragraphs, stories, etc.). 
Preliminary work in introducing planning modules in a hierarchical reinforcement learning (HRL) framework has shown a promising new direction for long form generation, specifically in video captioning~\cite{wang2018video}. 
With hierarchies, they were able to generate coherent expressions, offering richer semantic structures and better syntactic consistency. However, they train the Manager and Worker policies
alternately causing slow convergence issues for the HRL framework under the wake-sleep mode~\cite{Neal1996}. 

In visual storytelling, a latent world state is modeled by the words being generated. 
In this world, the scene in images, the players and the objects in it interact forming a coherent story of events.
A suitable generated story must be coherent at both the \textit{word-level}, linking words and phrases in a fluent way, and the \textit{world-level}, describing events that fulfill the underlying topics. Motivated by this observation, we design a network that generates a plan of sequence of topics for long-form generation. These topics are used to seed low-level text generator that produce words, conditioned on the intent of the topic vector. Rather than using an alternating policy training, we train the Manager and Worker \emph{jointly}, eliminating the problems of objective function balancing and slow convergence.

Building a sentence generator with a simple LSTM would not be sufficient to capture the linguistic features that are induced by the generated high level topics.  
A sentence generator that can provide an explicit semantic control over the sentence being generated can be better implemented with the semantic compositional network (SCN)~\cite{SCN_CVPR2017}, which we adopt in this work. 
Specifically, SCN adopts a mixture-of-expert design, which consists of a set of ``expert networks''. Each expert is itself an LSTM with specific parameters relating to the topics and combination of all expert output yields globally coherent sentence generation.

In the following, we first describe the encoder and the structure of the two-level decoders.
All $\Wmat$ matrices are projection matrices. For simplicity we omit the bias vectors from the formulation.

\subsection{Encoder}
In the visual story generation task, we are given a sequence of images \{$\mathbf{i}_1$,$\dots$,$\mathbf{i}_n$\} and a corresponding multi-sentence story in training. The image sequences are first embedded into image features $\{\vv_1,\ldots,\vv_n\}$ via a pre-trained CNN~\cite{he2016deep} and mean pooling is applied to generate an image-sequence content vector $\bar{\vv}$, which provides the whole model a global overview of the image-sequence content. This feature vector is then fed as initial input to the decoder. The words in the stories are embedded into word embedding vectors. At test time, the embeddings of the generated words are used.

\subsection{Two-Level Decoder}
Our two level Manager-Worker decoder is composed of two variants of LSTMs specifically designed for our topically structured network. 
\paragraph{Manager}
As shown in Fig.~\ref{fig:implementation_hrl_agent}, the Manager is implemented as an LSTM network, which uses the image-sequence content vector $\bar{\vv}$ as the initial input to the LSTM. 
The input to any time step of the $\ell$-th LSTM cell is the previous decoding output $\sv_{\ell-1}$, and the last hidden state $\hv_{\ell-1,T}$ ($T$-th output state) from the Worker LSTM (explained in detail later) after completing decoding of ($\ell-1$)-th sentence:
\begin{align}
	\sv_{\ell} &= \textrm{LSTM} (\sv_{\ell-1}, \hv_{\ell-1,T}) \\
	\cv_{\ell} &= [\vv_{\ell}, \sv_{\ell}] \\
	g_{\ell} &= \textrm{softmax} (\textrm{MLP}(\Wmat_1\cv_{\ell})) \,,
\end{align}
where $\textrm{MLP}[\cdot]$ denotes the multi-layer perception.
At each $\ell$-th time step of the Manager, the corresponding image feature $\vv_{\ell}$ is concatenated with the decoder output $\sv_{\ell}$, which acts as the context vector $\cv_{\ell}$ for the generation of sentence $\ell$ to fully describe the image content. Further, a new topic distribution $g_{\ell}$ is emitted via passing the resulting context vector $\cv_{\ell}$ through a softmax layer. 
The Manager decoder generates a topic distribution $g_{\ell}$ at each step of the decoder.
\paragraph{Worker}
The worker generates sentences given the context vector $\cv_{\ell}$ and the subgoal $\gv_{\ell}$ using semantic compositional network (SCN)~\cite{SCN_CVPR2017}, as illustrated in Fig.~\ref{fig:implementation_hrl_agent}. 

Specifically, we define two weight tensors $\mathbf{W}_3 \in \R^{n_h\times n_x \times K}$ and $\mathbf{W}_4 \in \R^{n_h\times n_h \times K}$, where $n_h$ is the number of hidden units, $n_x$ is the dimension of word embedding and $K$ is the number of topics. The $k$-th 2D ``slice'' of $\mathbf{W}_3[k]$ and $\mathbf{W}_4[k]$ represents parameters of the $k$-th expert. All $K$ experts work cooperatively to generate an output $a_{\ell,t}$:
\begin{align}
	a_{\ell,t} &= \textrm{softmax}(\textrm{MLP}(\Wmat_2\hv_{\ell,t}))  \label{eqn:scn_1} \\
	\hv_{\ell,t} &= \sigma(\Wmat_3(\gv_{\ell}) \xv_{\ell,t-1} + \Wmat_4(\gv_{\ell}) \hv_{\ell,t-1}) \,,  \label{eqn:scn_2}
\end{align} 
%
and

\begin{adjustbox}{minipage=1.08\linewidth,scale=0.93}
	\begin{align}
		\Wmat_3(\gv_{\ell}) =  \sum_{k=1}^K \gv_{\ell}[k]\Wmat_3[k], \,\, \Wmat_4(\gv_{\ell}) = \sum_{k=1}^K \gv_{\ell}[k] \Wmat_4[k] \label{eq:tensor}\,,
	\end{align}
\end{adjustbox}
where $\xv_{\ell,t-1}$ is the embedding of word $a_{\ell,t-1}$, and $\gv_{\ell}$ is a distributed vector with topic probabilities, rather than a one-hot vector. 

In order to reduce the number of model parameters, instead of implementing a tensor as in Eq.~(\ref{eq:tensor}), we decompose $\Wmat_3(\gv_{\ell})$:
\begin{align}
	\Wmat_3(\gv_{\ell}) = \Wmat_{3a} \cdot \text{diag}(\Wmat_{3b}\gv_{\ell}) \cdot \Wmat_{3c}
	\label{eqn:scn_3} 
\end{align}
into a multiplication of three terms $\Wmat_{3a}\in \R^{n_h\times n_f}$, $\Wmat_{3b}\in \R^{n_f\times T}$ and $\Wmat_{3c}\in \R^{n_f\times n_x}$, where $n_f$ is the number of factors.
The same factorization is also applied to $\Wmat_4(\gv_{\ell})$. 
$\Wmat_{3a}$ and $\Wmat_{3c}$ are shared by all the topics, while the diagonal term, $\textrm{diag}(\Wmat_{3b}\gv_{\ell})$, depends on the learned topics. Therefore, Eq.~(\ref{eqn:scn_3}) \emph{implicitly} defines $K$ LSTMs. Our Worker model can be considered as training an ensemble of up to $K$ LSTMs simultaneously.
When the Manager emits a subgoal to the Worker, it also \emph{implicitly} selects the corresponding LSTMs according to  topics' probabilities to generate the current sentence.

After a special end-of-sentence token is generated, the last hidden state $\hv_{\ell,T}$ of the Worker is sent to the Manager to emit the next subgoal $g_{\ell+1}$ distribution. 
This decoder defines the Worker policy,  which maps the current state to the vocabulary distribution given the goal distribution to sample the next word.
\subsection{Loss Functions}
Training a story generator using MLE produces stories that are locally coherent, but lack the topical content of the image thread. Using a scoring function that rewards the model for capturing story semantics, the model learns to produce generations that better represents the world state. We design loss functions to train the policies based on this goal. 

\paragraph{Manager Loss}
The Manager constructs a plan by generating a semantic concept (\emph{i.e.}, topic) for each image in sequence and is trained using MLE objective conditioned on the previous output and the current state information from the Worker $\hv_{\ell-1,T}$. It minimizes the negative log likelihood of predicting the next topic in the story given ground-truth topics $g_{\ell}^*$:
\begin{align}
	\Lcal_{\textrm{mle}}^M (\theta_m) = -\sum_{\ell=1}^{n} \log p_{\theta_m}(g_{\ell}^*|g_{1}^*,\ldots,g_{\ell-1}^*, h_{l-1,T}) \,,
	\label{managerpolicylearning}
\end{align}
where $\theta_m$ is the parameter vector of the Manager. This high-level semantic concept constrains the Worker's sentence generation policy. In the experiments, we define how we extract the ground-truth topic sequences $g_{1\dots n}^*$ from the stories. 
%
\paragraph{Worker Loss}
The Worker is responsible for generating sentences word by word. We define two different loss functions for the Worker training. The first is the MLE loss, which corresponds to the maximum-likelihood training: 
\begin{adjustbox}{minipage=1.10\linewidth,scale=0.91}
	\begin{align}
		\Lcal_{\textrm{mle}}^W (\theta_w)= -\sum_{\ell=1}^n \sum_{t=1}^{T} \log p_{\theta_w}(y_{\ell,t}^*|y_{\ell,1}^*,\ldots,y_{\ell,t-1}^*,g_{\ell},\cv_{\ell})\,,
		\label{eqn:mle_loss}
	\end{align} 
\end{adjustbox}
where $\theta_w$ is the parameter vector of the Worker; $y_{\ell}^*$ is the ground-truth sentence and $y_{\ell,t}^*$ denotes the $t$-th word in sentence $y_{\ell}^*$. The generation is conditioned on the goal distribution $g_{\ell}$ from the Manager and the context vector $\cv_{\ell}$.  

For the second loss, we assume the worker policy is stochastic and we learn it using the self-critical approach of~\cite{Rennie2016Self}. 
In self critical training, the model learns to gather more rewards from its generations by randomly sampling sequences that achieve higher reward than its best greedy samples. Two separate output sequences are sampled at each training iteration $t$: The first $\hat{y}$ is generated by randomly sampling from the model's distribution $p_{\theta_w}(\hat{y}_{\ell,t}|\hat{y}_{\ell,1},\ldots,\hat{y}_{\ell,t-1},g_{\ell},\cv_{\ell})$. The model's own outputs are the inputs at the next time step, acting similarly to test time conditions. Once the sentence is generated, a reward $\hat{r}$ is obtained. A second sequence $y^{\star}$, is generated by greedily sampling from the model's distribution $p_{\theta_w}(y^{\star}_{\ell,t}|y^{\star}_{\ell,1},\ldots,y^{\star}_{\ell,t-1},g_{\ell},\cv_{\ell})$ at each time step $t$ and a reward $r^{\star}$ is obtained. The following loss is used to train the self-critical RL method using the generated sequences and both rewards:

\begin{adjustbox}{minipage=1.10\linewidth,scale=0.88}
	\begin{align}
		\Lcal_{\textrm{rl}}^W (\theta_w) = -(r^{\star}-\hat{r}) \sum_{\ell=1}^n  \sum_{t=1}^{T}\log p_{\theta_w}(\hat{y}_{\ell,t}|\hat{y}_{\ell,1},\ldots,\hat{y}_{\ell,t-1},g_{\ell},\cv_{\ell})\,.
		\label{eqn:rl_loss}
	\end{align} 
\end{adjustbox}
The model encourages generating sequences that receive more reward than the best sequence that can be greedily sampled from the current policy. This way, self-critical training allows the model to explore sequences that yield higher reward than the current best policy. 
\paragraph{Mixed Worker Loss}
Minimizing the RL loss in Eq.~(\ref{eqn:rl_loss}) alone does not ensure the readability and fluency of the generated sentences. The model quickly learns to generate simple sequences that \textit{exploit} the teacher for higher rewards despite producing nonsensical sentences. To remedy this, a better way is to optimize a mixed objective~\cite{wu2016google,Pasunuru2017Reinforced} that balances learning for generating coherent story sentences with maintaining generator's language model:
\begin{align}
	\Lcal_{\textrm{mix}}^W = \gamma \Lcal_{\textrm{rl}}^W  +(1-\gamma) \Lcal_{\textrm{mle}}^W \,,
	\label{eqn:mixedLoss}
\end{align}
where $\gamma \in [0,1]$ is a scaling factor balancing the importance of $\Lcal_{\textrm{rl}}^W$ and $\Lcal_{\textrm{mle}}^W$. 
For annealing and faster convergence, we start with $\gamma=0$ (\emph{i.e.}, minimizing the cross-entropy loss), and gradually increase $\gamma$ to a maximum value $\gamma_{max}<1$.
\subsection{Policy Learning} \label{subsec:End-to-EndTraining}
We investigate 3 objectives to combine the Manager and Worker policies in an end-to-end learning framework. 

\paragraph{Cascaded Training}
The Manager and the Worker are trained independently. The manager is trained using the MLE loss $\Lcal_{\textrm{mle}}^M$ = $-\sum_{\ell=1}^{n} \log p(g_{\ell}^*|g_{1}^*,\ldots,g_{\ell-1}^*)$ without any input from the Worker. Once the Manager training is converged, the Worker uses the trained Manager model to generate a topic sequence given each image sequence. 
The Worker is trained using the mixed loss $\Lcal_{\textrm{mix}}^W$ with the ground-truth topic sequence. 

\paragraph{Iterative Training (Wake-Sleep Mode)}
The Manager and Worker are trained iteratively in a wake-sleep mode similar to HRL training of \cite{wang2018video}.
The Manager observes the current state $\hv_{\ell-1,T}$ after the Worker generates sentence $\ell-1$, and produces a topic distribution $g_{\ell}$ for the generation of sentence $\ell$. The Worker takes as input a state $\hv_{\ell,t-1}$ and a topic distribution $g_{\ell}$, and predicts its next action $a_{\ell,t}$, \emph{i.e.}, the $t$-th word in sentence $\ell$. Note that the the topic distribution $g_{\ell}$ at Manager decoder time $\ell$ serves as a guidance and remains a constant input to the Worker decoder during the whole process of generating sentence $\ell$. 

In the early stage of training, we set $\gamma=0$ to pretrain the Worker policy with $\Lcal_{\textrm{mle}}^W$. This ensures that our Worker RL agent training starts at a good initialization point. 
After the warm-up pre-training, the Worker policy and the Manager policy are trained iteratively while keeping the other one fixed, using the losses $\Lcal_{\textrm{mle}}^M$ and $\Lcal_{\textrm{mix}}^W$, respectively. In training the Worker, a sentence-level CIDEr score~\cite{vedantam2015cider} is used as the intermediate reward, which measures how well a local sentence matches the ground-truth. 

\paragraph{Joint Training}
Iterative training may yield instability and slow convergence issues in objective function optimization. An alternative is to use a joint training scheme to enable the Manager and Worker to backpropogate from each other's losses and optimize globally. We introduce a new joint training objective combining the Manager and the Worker losses:
\begin{align}
	\Lcal_{\textrm{joint}} = (1-\gamma_1) \Lcal_{\textrm{mle}}^M + \gamma_1(\gamma_2 \Lcal_{\textrm{rl}}^W  +(1-\gamma_2) \Lcal_{\textrm{mle}}^W)\,.
	\label{jointt}
\end{align}
Similar to iterative training, in the early stages of training, we set $\gamma_2=0$ to pretrain the Worker policy with MLE loss. 
After the warm-up pre-training, the Worker policy and the Manager policy are trained jointly using Eq.~(\ref{jointt}). Here, we also use sentence-level CIDEr score as the intermediate reward. While the Worker aims to learn a better sentence generator, the Manager aims to learn a better paragraph planner by optimizing for high-level topic semantics. The Manager's parameters are updated based on the rewards the Worker receives upon generating a story.  
\paragraph{Sentence Level Credit Assignment}
The Worker takes actions by generating words until a special end-of-sentence token is reached, then it obtains a reward. We evaluate the model on story level until all sentences are generated.  
In multi-sentence generation tasks, the final reward after full story is generated can be a weak signal. To alleviate that, we also use \emph{intermediate} rewards by assigning sentence-level rewards, by evaluating how well the current generated sentence matches the ground-truth. 
This intermediate reward helps alleviate the reward sparsity problem, and in experiments, we found that it also helps reduce sentence and word repetition issues, yielding more diverse stories.

\section{Experimental Results}
\label{sec:ExperimentalResults}
\paragraph{Dataset}\label{subsec:Dataset}
For learning and evaluation we use the VIST dataset~\cite{huang2016visual}, which are collected from Flickr albums and then annotated by Amazon's Mechanical Turk (AMT). Each story has 5 images and 5 corresponding descriptions. After filtering out broken images, we obtain 19,828 image sequences with 49,629 stories in total. On average, each image sequence is annotated with roughly 2.5 stories. The 19,828 image sequences are partitioned into three parts, 15,851 for training, 1,976 for validation and 2,001 for testing, respectively. Correspondingly, the 49,629 stories are also split into three parts, 39,676 for training, 4,943 for validation and 5,010 for testing, respectively. The vocabulary consists of 12,977 words. 

\paragraph{Training}\label{subsec:Training}
We extract the image features with ResNet-152~\cite{he2016deep} pretrained on the
ImageNet dataset. The resulting image feature vector $\vv$ has 2,048 dimensions.
We use the GLove embedding vectors of \cite{pennington2014glove} for word embedding initialization.

Since the VIST dataset \cite{huang2016visual} is originally not annotated with topic sequences, we use clustering to generate golden topic sequences. Specifically, we use a simple k-means algorithm to cluster the ResNet-152 image features into $K$ clusters, where each cluster implicitly defines a topic, and the  sentences are then considered as belonging to the same cluster as the corresponding images. 


{\small
	\begin{table*}[htb]
		\small
		\caption{\small Evaluation results for generated stories by models and baselines. \textbf{bold} the top performing result. The Worker+Random topics and Worker+GTT are the lower and upper bound scores for our Hierarchically Structured RL (HSRL) model. }
		\label{table:BLEU}
		\centering
		\begin{tabular}{lcccccc}
			\hline
			Methods       & BLEU-4 & ROUGE-L &   CIDEr-D & METEOR-v1 & METEOR-v2 & SPICE\\
			\hline
			seq2seq + heuristics~\cite{huang2016visual} & 3.50 & $-$ & 6.84 & 10.25 & 31.4 & $-$\\
			BARNN~\cite{liu2017let} & $-$ & $-$ & $-$ & $-$ & 33.3 & $-$\\
			h-attn-rank~\cite{yu2017hierarchically} & $-$ & 29.8 & 7.38  & $-$ & 33.9 & $-$\\
			AREL~\cite{wang2018no} & \textbf{14.1} & 29.6 & 9.5  & $-$ & 35.2 & $-$\\
			Show, Reward \& Tell~\cite{wang2018show} & 5.16 & $-$ & \textbf{11.35}  & 12.32 & $-$ & $-$\\
			\hline
			\multicolumn{7}{l}{\textbf{Our Baselines}} \\
			\hline
			Baseline LSTM (MLE)  & 7.32 & 27.34 & 7.52 & 8.04 & 31.43 & 7.03\\
			Baseline LSTM (RL)  & 8.16 & 27.52 & 7.64 & 8.31 & 31.52 & 7.57\\
			HRL~\cite{wang2018video} & 8.94 & 27.90 & 8.72 & 11.4 & 32.67 & 8.73 \\
			\hline
			\multicolumn{7}{l}{\textbf{16 Topics}} \\
			\hline
			Worker+Random topics & 4.70 & 23.04 & 3.90 & 5.43 & 27.11 & 5.54\\
			HSRL w/ Cascaded Training & 10.38 & 30.14 & 9.65 & 12.32 & 34.73 & 9.62\\
			HSRL w/ Iterative Training & 11.23 & 30.32 & 9.68 & 12.83 & 34.82 & 9.87 \\
			HSRL w/ Joint Training & \textbf{11.64} & \textbf{30.61} &  \textbf{9.73}&  \textbf{13.27}&  \textbf{34.95} &   \textbf{10.25} \\
			Worker+GTT & 13.41 & 31.53 & 10.82 & 14.27 & 35.48 & 12.83 \\
			\hline
			\multicolumn{7}{l}{\textbf{64 Topics}} \\
			\hline
			HSRL w/ Cascaded Training & 11.95 & 30.06 & 10.03 & 13.34 & 34.81 & 12.42 \\
			HSRL w/ Iterative Training & 12.04 & 30.65 & 10.34 & 13.42 & 35.21 & 12.66 \\
			HSRL w/ Joint Training & \textbf{12.32} & \textbf{30.84}&  \textbf{10.71} &  \textbf{13.53} & \textbf{35.23} &   \textbf{12.97}\\
			Worker+GTT & 14.68 & 32.73 & 12.63 & 16.32 & 36.22 & 14.34 \\
			\hline
		\end{tabular}
	\end{table*}
}

\subsection{Results}\label{subsec:Evaluation}
\paragraph{Scores} 
We compute BLEU-4~\cite{papineni2002bleu}, METEOR~\cite{banerjee2005meteor}, CIDEr~\cite{vedantam2015cider}, ROUGE-L~\cite{lin2004rouge}, and SPICE~\cite{anderson2016spice} metrics for evaluation. 

\paragraph{Baselines} 
We provide results reported in previous methods: \cite{huang2016visual} adds a decoder-time heuristic method to alleviate the repetition issue when generating stories, \cite{liu2017let} uses an additional cross-modality embedding model to regularize the story generation model, while \cite{yu2017hierarchically} uses a hierarchical model, and considers performing album summarization and storytelling simultaneously. All these models are trained using the MLE loss. Recently,~\cite{wang2018no} and~\cite{wang2018show} proposes the usage of a learned reward in an RL setup for improving the performance. Our model also uses RL, but with a different focus on using learned topics for effective paragraph planning. 

Note that we observe some discrepancy in the reported results of the related work (see Table~\ref{table:BLEU}). Specifically comparing to closest works to ours, in ~\cite{wang2018show}, a high CIDEr-D and a low BLEU-4 score is reported, while in ~\cite{wang2018no} even though a much higher BLEU-4 score is reported, their CIDEr-D score is much lower. 
These discrepancies are possibly due to differences in data pre-processing and evaluation scripts. 

Therefore, for fair comparison, we mainly focus on comparing our results with the following re-implemented baselines: Baseline LSTM (MLE) is trained with cross-entropy loss using Eq.~(\ref{eqn:mle_loss}), while Baseline LSTM (RL) is trained with self-critical REINFORCE loss using Eq.~(\ref{eqn:rl_loss}). We re-implemented the HRL approach in~\cite{wang2018video} for our task, and also implemented a variant of our model, Worker+Random topics, which learns a Worker decoder without the Manager decoder but randomly samples a sequence of topics from an interval of [1,$K$], where $K$ is the total number of topics. We use this baseline as a lower-bound of our model. 
Similarly, we also implemented a variant of our model, Worker+GTT, again with only a Worker decoder, but this time we used the ground-truth topics (GTT) as input to the Worker. We use this baseline as an upper-bound of our model. We experimented with $K$=16 and $K$=64. 
All the metrics are computed by using the code released by the COCO evaluation server~\cite{chen2015microsoft}.

\begin{figure*}[t]
	\centering
	\includegraphics[width=0.92\textwidth]{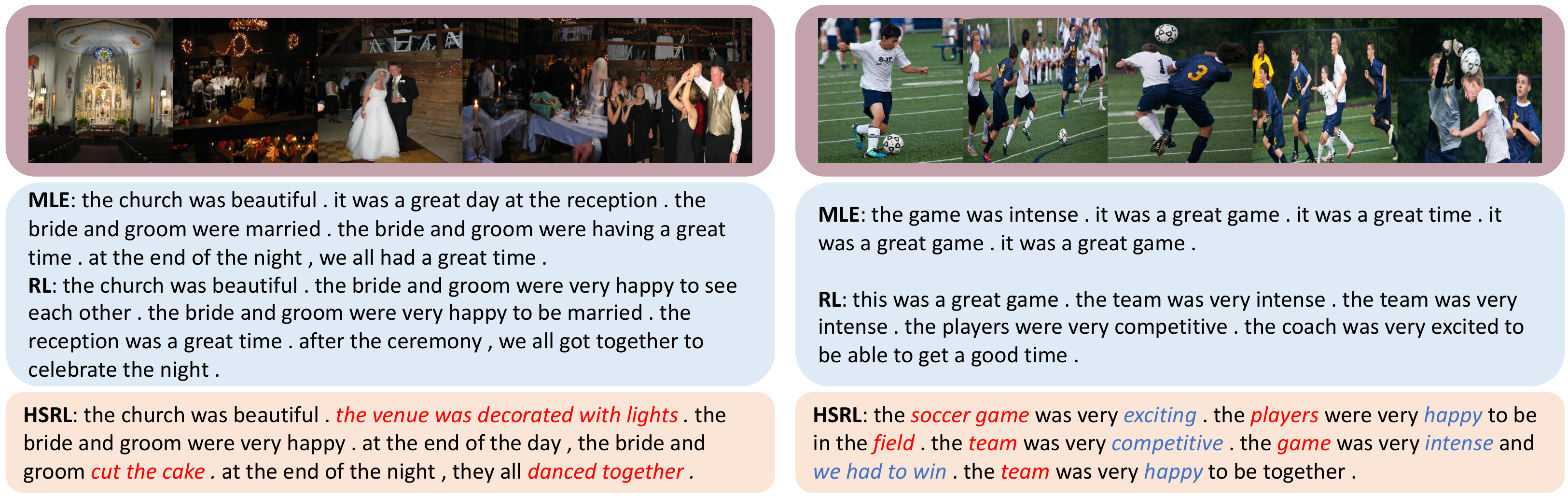}
	\caption{\small Example stories generated by three storytelling models. Compared to baseline models, the hierarchically structured model generates more coherent, detailed and expressive paragraphs.}
	\label{fig:stories}
\end{figure*}

\paragraph{Quantitative Results} 
Our results\footnote{METEOR-v1 represents the version used in the COCO evaluation server, and METEOR-v2 represents version 1.5 with \texttt{HTER} weights.} in Table~\ref{table:BLEU} show that models optimized with joint training achieve the greatest improvement for all the scores.
All hierarchically structured reinforced (HSRL) story generation models outperform the baseline LSTM (MLE) and LSTM (RL) training models by a margin. HRL only improves flat RL marginally in this task, while our HSRL achieved much better performance than HRL.
This indicates the efficiency of using explicit topics as subgoals, rather than a latent continuous vector as used in~\cite{wang2018video}.
Additionally, HSRL models consistently achieve high improvements across different number of topics against the lower-bound Worker+Random model, which was trained using random topic sequences. The joint training results are close to the upper bound Worker+GTT model, indicating the stronger performance of joint training.

Careful analysis of these three end-to-end training methods yields that the model optimized with cascaded training method performs worse than the rest of the training methods. Iterative training with HSRL improves the results over the cascaded training method across all scores,  
indicating the impact of sentence planning with higher-level decoder for representing higher level semantics of the story generation. The model trained jointly with HSRL achieves even higher scores, across different numbers of topics, showing the benefits of training a two-level decoder network jointly rather than iteratively in an alternating training mode. The success of joint training can also be attributed to the fact that at training time, both Manager and Worker have access to the ground truth topic and word sequences through reward functions and teacher forcing~\cite{lamb2016professor} in MLE training, while in iterative training the Manager and Worker has access to either topics or word sequences. 

\paragraph{Human Evaluation}
We perform two human evaluation tasks using Amazon Mechanical Turk: pairwise comparison and the closeness to the ground-truth story.  
For both tasks, we use all the 2001 test image sequences. Each image sequence is presented to the worker and the worker is requested to judge which generated story is better in terms of relevance, expressiveness and concreteness. A neutral option is provided if the worker cannot tell which one is better. The compared approaches include MLE, RL and our HSRL model. Results are summarized in Table~\ref{tbl:human_pair}. It is clearly shown that our HSRL approach is better than MLE and RL. 

The second task requires the AMT-worker to judge the closeness of each story to the ground-truth story comparing against the other two generated stories.  
HSRL wins 55.87\% tasks, RL (the second best ) wins 34.53\%, while MLE only wins 9.60\%. 

\paragraph{Qualitative Analysis} In Fig.~\ref{fig:framework} we illustrate the paragraph generation process by explicitly showing the generated topic sequences. A high-level plan is first constructed by generating a sequence of topics, based on which a sequence of sentences are then generated. Specifically in this example, our model constructs a plan as follow: (\emph{i}) describe the story background; (\emph{ii}) describe the little girl, family and baby sequentially; (\emph{iii}) end the story with ``everyone had a great time''.

In Fig.~\ref{fig:stories}, we show two additional examples that are sampled from the test set. We compare our HSRL model with two of our baseline models and obtain the following observations. First, our HSRL model is globally coherent. It can describe images in connection to previous images more coherently. For instance, in Fig.~\ref{fig:stories}(right), the model identifies that the setting of the story is \textit{soccer game} and follows to explain that there were \textit{players}, \textit{field} and \textit{team} in the scene, in accordance of their appearance in image sequences.

The generated stories also indicate that the sentences generated by HSRL is more diverse compared to the baselines. 
For instance each sentence generated by HSRL in Fig.~\ref{fig:stories}(right) is different from others, which can be attributed to the fact that sentence generation is nicely controlled by the topics generated for the story. In contrast, the stories generated by two baseline models are less diverse and contains repetitions. As a by-product, it was exciting to observe that the stories generated by the HSRL are more vibrant and engaging. Just like a human-created story that touches on human emotions, we found that the story generated by the HSRL is more emotional. For example, the words ``\textit{happy, exciting/excited, fun/funny, intense, tired}'' have been used 1137, 678, 316, 51, and 58 times in HSRL generated stories, and 410, 138, 291, 38 and 3 times in the RL baseline.

Finally, we also observe that the HSRL model was able to learn to exploit the reward function to include more details in the generated text. For instance, in Fig.~\ref{fig:stories}(left), though stories generated by all the 3 models are reasonable, we observe less details in the stories generated by the baselines, while the salient facts like ``decorated with lights, cut the cake, danced together'' are captured by our model. More examples are provided in the supplementary material.

\begin{table}[t!]
	\centering
	\caption{\small Results of pairwise human comparison. }
	\label{tbl:human_pair}
	\scriptsize
	\begin{tabular}{c|ccc|ccc}
		\hline
		& \multicolumn{3}{c|}{MLE vs HSRL} & \multicolumn{3}{c}{RL vs HSRL} \\
		\hline
		Choice (\%) & MLE & HSRL & Tie & RL & HSRL & Tie \\
		\hline
		Relevance & 27.53 & 63.93 & 8.53 & 34.87 & 56.40 & 8.73 \\ 
		Expressiveness & 24.87 & 62.53 & 12.60 & 31.60 & 55.67 & 12.73 \\ 
		Concreteness & 25.87 & 62.47 & 11.66 & 33.93 & 54.73 & 11.33 \\
		\hline
	\end{tabular}
\end{table} 




\section{Conclusion} \label{sec:Conclusion}
We investigated the problem of generating topically coherent visual stories given an image stream and demonstrated that the use of hierarchically structured reinforcement learning can improve the generation. Analysis demonstrates that this improvement is due to the joint training of two hierarchically structured decoders, where the higher decoder is optimized for better learning high-level topical semantics, and the lower decoder optimizes to obtain more rewards for generating topically coherent sentences.  

\paragraph{Acknowledgements} We thank Xiujun Li and Chuang Gan for helpful discussions. We thank anonymous reviewers for their constructive feedbacks. 

\small{
	\bibliographystyle{aaai}
	\bibliography{reference}

\begin{thebibliography}{}

\bibitem[\protect\citeauthoryear{Anderson \bgroup et al\mbox.\egroup
  }{2016}]{anderson2016spice}
Anderson, P.; Fernando, B.; Johnson, M.; and Gould, S.
\newblock 2016.
\newblock Spice: Semantic propositional image caption evaluation.
\newblock In {\em ECCV}.

\bibitem[\protect\citeauthoryear{Banerjee and Lavie}{2005}]{banerjee2005meteor}
Banerjee, S., and Lavie, A.
\newblock 2005.
\newblock Meteor: An automatic metric for mt evaluation with improved
  correlation with human judgments.
\newblock In {\em ACL Workshop}.

\bibitem[\protect\citeauthoryear{Bengio \bgroup et al\mbox.\egroup
  }{2015}]{bengio2015scheduled}
Bengio, S.; Vinyals, O.; Jaitly, N.; and Shazeer, N.
\newblock 2015.
\newblock Scheduled sampling for sequence prediction with recurrent neural
  networks.
\newblock In {\em NIPS}.

\bibitem[\protect\citeauthoryear{Celikyilmaz \bgroup et al\mbox.\egroup
  }{2018}]{Celikyilmaz2018DCA}
Celikyilmaz, A.; Bosselut, A.; He, X.; and Choi, Y.
\newblock 2018.
\newblock Deep communicating agents for abstractive summarization.
\newblock In {\em NAACL}.

\bibitem[\protect\citeauthoryear{Chen \bgroup et al\mbox.\egroup
  }{2015}]{chen2015microsoft}
Chen, X.; Fang, H.; Lin, T.-Y.; Vedantam, R.; Gupta, S.; Doll{\'a}r, P.; and
  Zitnick, C.~L.
\newblock 2015.
\newblock Microsoft coco captions: Data collection and evaluation server.
\newblock {\em arXiv preprint arXiv:1504.00325}.

\bibitem[\protect\citeauthoryear{Cho \bgroup et al\mbox.\egroup
  }{2014}]{cho2014learning}
Cho, K.; Van~Merri{\"e}nboer, B.; Gulcehre, C.; Bahdanau, D.; Bougares, F.;
  Schwenk, H.; and Bengio, Y.
\newblock 2014.
\newblock Learning phrase representations using rnn encoder-decoder for
  statistical machine translation.
\newblock In {\em EMNLP}.

\bibitem[\protect\citeauthoryear{Dayan and Neil}{1996}]{Neal1996}
Dayan, P., and Neil, R.~M.
\newblock 1996.
\newblock Factor analysis using delta-rule wake-sleep learning.
\newblock {\em Technical Report, Department of Statistics, University of
  Toronto}.

\bibitem[\protect\citeauthoryear{Donahue \bgroup et al\mbox.\egroup
  }{2015}]{donahue2015long}
Donahue, J.; Anne~Hendricks, L.; Guadarrama, S.; Rohrbach, M.; Venugopalan, S.;
  Saenko, K.; and Darrell, T.
\newblock 2015.
\newblock Long-term recurrent convolutional networks for visual recognition and
  description.
\newblock In {\em CVPR}.

\bibitem[\protect\citeauthoryear{Fang \bgroup et al\mbox.\egroup
  }{2015}]{fang2015captions}
Fang, H.; Gupta, S.; Iandola, F.; Srivastava, R.~K.; Deng, L.; Doll{\'a}r, P.;
  Gao, J.; He, X.; Mitchell, M.; Platt, J.~C.; et~al.
\newblock 2015.
\newblock From captions to visual concepts and back.
\newblock In {\em CVPR}.

\bibitem[\protect\citeauthoryear{Gan \bgroup et al\mbox.\egroup
  }{2017}]{SCN_CVPR2017}
Gan, Z.; Gan, C.; He, X.; Pu, Y.; Tran, K.; Gao, J.; Carin, L.; and Deng, L.
\newblock 2017.
\newblock Semantic compositional networks for visual captioning.
\newblock In {\em CVPR}.

\bibitem[\protect\citeauthoryear{He \bgroup et al\mbox.\egroup
  }{2016}]{he2016deep}
He, K.; Zhang, X.; Ren, S.; and Sun, J.
\newblock 2016.
\newblock Deep residual learning for image recognition.
\newblock In {\em CVPR}.

\bibitem[\protect\citeauthoryear{Hochreiter and
  Schmidhuber}{1997}]{hochreiter1997long}
Hochreiter, S., and Schmidhuber, J.
\newblock 1997.
\newblock Long short-term memory.
\newblock {\em Neural computation}.

\bibitem[\protect\citeauthoryear{Huang \bgroup et al\mbox.\egroup
  }{2016}]{huang2016visual}
Huang, T.-H.~K.; Ferraro, F.; Mostafazadeh, N.; Misra, I.; Agrawal, A.; Devlin,
  J.; Girshick, R.; He, X.; Kohli, P.; Batra, D.; et~al.
\newblock 2016.
\newblock Visual storytelling.
\newblock In {\em NAACL}.

\bibitem[\protect\citeauthoryear{Karpathy and Fei-Fei}{2015}]{karpathy2015deep}
Karpathy, A., and Fei-Fei, L.
\newblock 2015.
\newblock Deep visual-semantic alignments for generating image descriptions.
\newblock In {\em CVPR}.

\bibitem[\protect\citeauthoryear{Krause \bgroup et al\mbox.\egroup
  }{2017}]{krause2016hierarchical}
Krause, J.; Johnson, J.; Krishna, R.; and Fei-Fei, L.
\newblock 2017.
\newblock A hierarchical approach for generating descriptive image paragraphs.
\newblock In {\em CVPR}.

\bibitem[\protect\citeauthoryear{Krishna \bgroup et al\mbox.\egroup
  }{2017}]{krishna2017dense}
Krishna, R.; Hata, K.; Ren, F.; Fei-Fei, L.; and Niebles, J.~C.
\newblock 2017.
\newblock Dense-captioning events in videos.
\newblock In {\em ICCV}.

\bibitem[\protect\citeauthoryear{Lamb \bgroup et al\mbox.\egroup
  }{2016}]{lamb2016professor}
Lamb, A.~M.; GOYAL, A. G. A.~P.; Zhang, Y.; Zhang, S.; Courville, A.~C.; and
  Bengio, Y.
\newblock 2016.
\newblock Professor forcing: A new algorithm for training recurrent networks.
\newblock In {\em NIPS}.

\bibitem[\protect\citeauthoryear{Lin}{2004}]{lin2004rouge}
Lin, C.-Y.
\newblock 2004.
\newblock Rouge: A package for automatic evaluation of summaries.
\newblock In {\em ACL Workshop}.

\bibitem[\protect\citeauthoryear{Liu \bgroup et al\mbox.\egroup
  }{2017a}]{liu2016improved}
Liu, S.; Zhu, Z.; Ye, N.; Guadarrama, S.; and Murphy, K.
\newblock 2017a.
\newblock Improved image captioning via policy gradient optimization of spider.
\newblock In {\em ICCV}.

\bibitem[\protect\citeauthoryear{Liu \bgroup et al\mbox.\egroup
  }{2017b}]{liu2017let}
Liu, Y.; Fu, J.; Mei, T.; and Chen, C.~W.
\newblock 2017b.
\newblock Let your photos talk: Generating narrative paragraph for photo stream
  via bidirectional attention recurrent neural networks.
\newblock In {\em AAAI}.

\bibitem[\protect\citeauthoryear{Pan \bgroup et al\mbox.\egroup
  }{2016a}]{pan2015hierarchical}
Pan, P.; Xu, Z.; Yang, Y.; Wu, F.; and Zhuang, Y.
\newblock 2016a.
\newblock Hierarchical recurrent neural encoder for video representation with
  application to captioning.
\newblock In {\em CVPR}.

\bibitem[\protect\citeauthoryear{Pan \bgroup et al\mbox.\egroup
  }{2016b}]{pan2016joint}
Pan, Y.; Mei, T.; Yao, T.; Li, H.; and Rui, Y.
\newblock 2016b.
\newblock Jointly modeling embedding and translation to bridge video and
  language.
\newblock In {\em CVPR}.

\bibitem[\protect\citeauthoryear{Papineni \bgroup et al\mbox.\egroup
  }{2002}]{papineni2002bleu}
Papineni, K.; Roukos, S.; Ward, T.; and Zhu, W.-J.
\newblock 2002.
\newblock Bleu: a method for automatic evaluation of machine translation.
\newblock In {\em ACL}.

\bibitem[\protect\citeauthoryear{Park and Kim}{2015}]{park2015expressing}
Park, C.~C., and Kim, G.
\newblock 2015.
\newblock Expressing an image stream with a sequence of natural sentences.
\newblock In {\em NIPS}.

\bibitem[\protect\citeauthoryear{Pasunuru and
  Bansal}{2017}]{Pasunuru2017Reinforced}
Pasunuru, R., and Bansal, M.
\newblock 2017.
\newblock Reinforced video captioning with entailment rewards.
\newblock In {\em EMNLP}.

\bibitem[\protect\citeauthoryear{Paulus, Xiong, and
  Socher}{2018}]{paulus2017deep}
Paulus, R.; Xiong, C.; and Socher, R.
\newblock 2018.
\newblock A deep reinforced model for abstractive summarization.
\newblock In {\em ICLR}.

\bibitem[\protect\citeauthoryear{Pennington, Socher, and
  Manning}{2014}]{pennington2014glove}
Pennington, J.; Socher, R.; and Manning, C.
\newblock 2014.
\newblock Glove: Global vectors for word representation.
\newblock In {\em EMNLP}.

\bibitem[\protect\citeauthoryear{Pu \bgroup et al\mbox.\egroup
  }{2018}]{pu2018adaptive}
Pu, Y.; Min, M.~R.; Gan, Z.; and Carin, L.
\newblock 2018.
\newblock Adaptive feature abstraction for translating video to text.
\newblock In {\em AAAI}.

\bibitem[\protect\citeauthoryear{Ranzato \bgroup et al\mbox.\egroup
  }{2016}]{ranzato2015sequence}
Ranzato, M.; Chopra, S.; Auli, M.; and Zaremba, W.
\newblock 2016.
\newblock Sequence level training with recurrent neural networks.
\newblock In {\em ICLR}.

\bibitem[\protect\citeauthoryear{Ren \bgroup et al\mbox.\egroup
  }{2017}]{ren2017deep}
Ren, Z.; Wang, X.; Zhang, N.; Lv, X.; and Li, L.-J.
\newblock 2017.
\newblock Deep reinforcement learning-based image captioning with embedding
  reward.
\newblock In {\em CVPR}.

\bibitem[\protect\citeauthoryear{Rennie \bgroup et al\mbox.\egroup
  }{2017}]{Rennie2016Self}
Rennie, S.~J.; Marcheret, E.; Mroueh, Y.; Ross, J.; and Goel, V.
\newblock 2017.
\newblock Self-critical sequence training for image captioning.
\newblock In {\em CVPR}.

\bibitem[\protect\citeauthoryear{Sutskever, Vinyals, and
  Le}{2014}]{sutskever2014sequence}
Sutskever, I.; Vinyals, O.; and Le, Q.~V.
\newblock 2014.
\newblock Sequence to sequence learning with neural networks.
\newblock In {\em NIPS}.

\bibitem[\protect\citeauthoryear{Vedantam, Lawrence~Zitnick, and
  Parikh}{2015}]{vedantam2015cider}
Vedantam, R.; Lawrence~Zitnick, C.; and Parikh, D.
\newblock 2015.
\newblock Cider: Consensus-based image description evaluation.
\newblock In {\em CVPR}.

\bibitem[\protect\citeauthoryear{Venugopalan \bgroup et al\mbox.\egroup
  }{2015}]{venugopalan2015sequence}
Venugopalan, S.; Rohrbach, M.; Donahue, J.; Mooney, R.; Darrell, T.; and
  Saenko, K.
\newblock 2015.
\newblock Sequence to sequence-video to text.
\newblock In {\em ICCV}.

\bibitem[\protect\citeauthoryear{Vinyals \bgroup et al\mbox.\egroup
  }{2015}]{vinyals2015show}
Vinyals, O.; Toshev, A.; Bengio, S.; and Erhan, D.
\newblock 2015.
\newblock Show and tell: A neural image caption generator.
\newblock In {\em CVPR}.

\bibitem[\protect\citeauthoryear{Wang \bgroup et al\mbox.\egroup
  }{2018a}]{wang2018show}
Wang, J.; Fu, J.; Tang, J.; Li, Z.; and Mei, T.
\newblock 2018a.
\newblock Show, reward and tell: Automatic generation of narrative paragraph
  from photo stream by adversarial training.
\newblock In {\em AAAI}.

\bibitem[\protect\citeauthoryear{Wang \bgroup et al\mbox.\egroup
  }{2018b}]{wang2018no}
Wang, X.; Chen, W.; Wang, Y.-F.; and Wang, W.~Y.
\newblock 2018b.
\newblock No metrics are perfect: Adversarial reward learning for visual
  storytelling.
\newblock In {\em ACL}.

\bibitem[\protect\citeauthoryear{Wang \bgroup et al\mbox.\egroup
  }{2018c}]{wang2018video}
Wang, X.; Chen, W.; Wu, J.; Wang, Y.-F.; and Wang, W.~Y.
\newblock 2018c.
\newblock Video captioning via hierarchical reinforcement learning.
\newblock In {\em CVPR}.

\bibitem[\protect\citeauthoryear{Williams}{1992}]{williams1992simple}
Williams, R.~J.
\newblock 1992.
\newblock Simple statistical gradient-following algorithms for connectionist
  reinforcement learning.
\newblock {\em Reinforcement Learning}.

\bibitem[\protect\citeauthoryear{Wu \bgroup et al\mbox.\egroup
  }{2016}]{wu2016google}
Wu, Y.; Schuster, M.; Chen, Z.; Le, Q.~V.; Norouzi, M.; Macherey, W.; Krikun,
  M.; Cao, Y.; Gao, Q.; Macherey, K.; et~al.
\newblock 2016.
\newblock Google's neural machine translation system: Bridging the gap between
  human and machine translation.
\newblock {\em arXiv preprint arXiv:1609.08144}.

\bibitem[\protect\citeauthoryear{Xu \bgroup et al\mbox.\egroup
  }{2015}]{xu2015show}
Xu, K.; Ba, J.; Kiros, R.; Cho, K.; Courville, A.; Salakhutdinov, R.; Zemel,
  R.~S.; and Bengio, Y.
\newblock 2015.
\newblock Show, attend and tell: Neural image caption generation with visual
  attention.
\newblock In {\em ICML}.

\bibitem[\protect\citeauthoryear{Yu, Bansal, and
  Berg}{2017}]{yu2017hierarchically}
Yu, L.; Bansal, M.; and Berg, T.~L.
\newblock 2017.
\newblock Hierarchically-attentive rnn for album summarization and
  storytelling.
\newblock In {\em EMNLP}.

\bibitem[\protect\citeauthoryear{Yu \bgroup et al\mbox.\egroup
  }{2016}]{yu2016video}
Yu, H.; Wang, J.; Huang, Z.; Yang, Y.; and Xu, W.
\newblock 2016.
\newblock Video paragraph captioning using hierarchical recurrent neural
  networks.
\newblock In {\em CVPR}.

\end{thebibliography}
}

\clearpage
\twocolumn[{%
	\centering
	\Large \bf{Supplementary Material for ``Hierarchically Structured Reinforcement Learning for \\ Topically Coherent Visual Story Generation'' }
	\\[1.5em]
	\section{More Examples}
	\begin{center}
		\centering
		\includegraphics[width=0.92\textwidth]{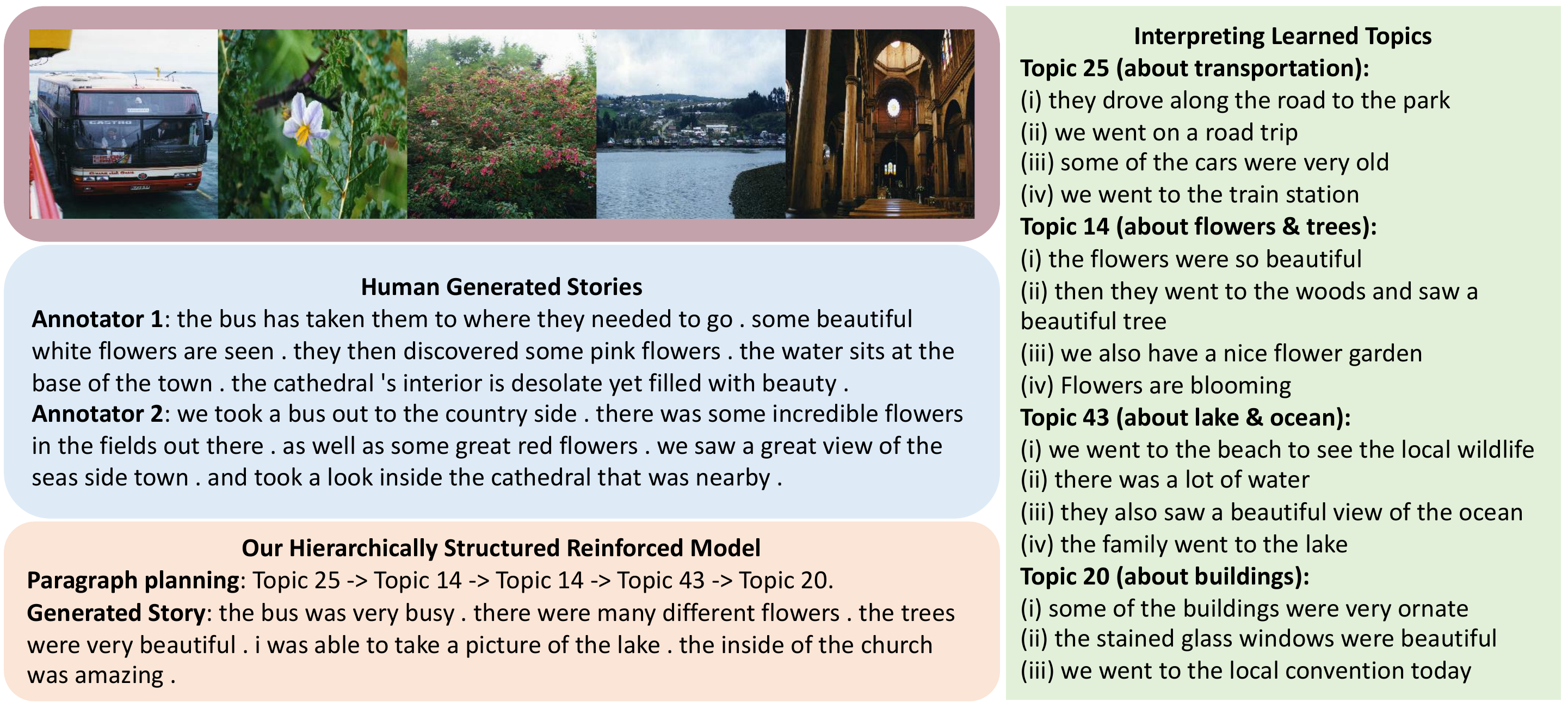}
		\captionof{figure}{Another example of HSRL-based visual storytelling. Our model generates coherent stories by paragraph planning, \emph{i.e.}, predicting a sequence of topics. In order to visualize learned topics, we present sentences generated from the corresponding topics in the test set. We manually assigned the topic names in this example for visual clarity.
			\small }
		\label{fig:example_more}
	\end{center}
	\begin{center}
		\centering
		\includegraphics[width=0.92\textwidth]{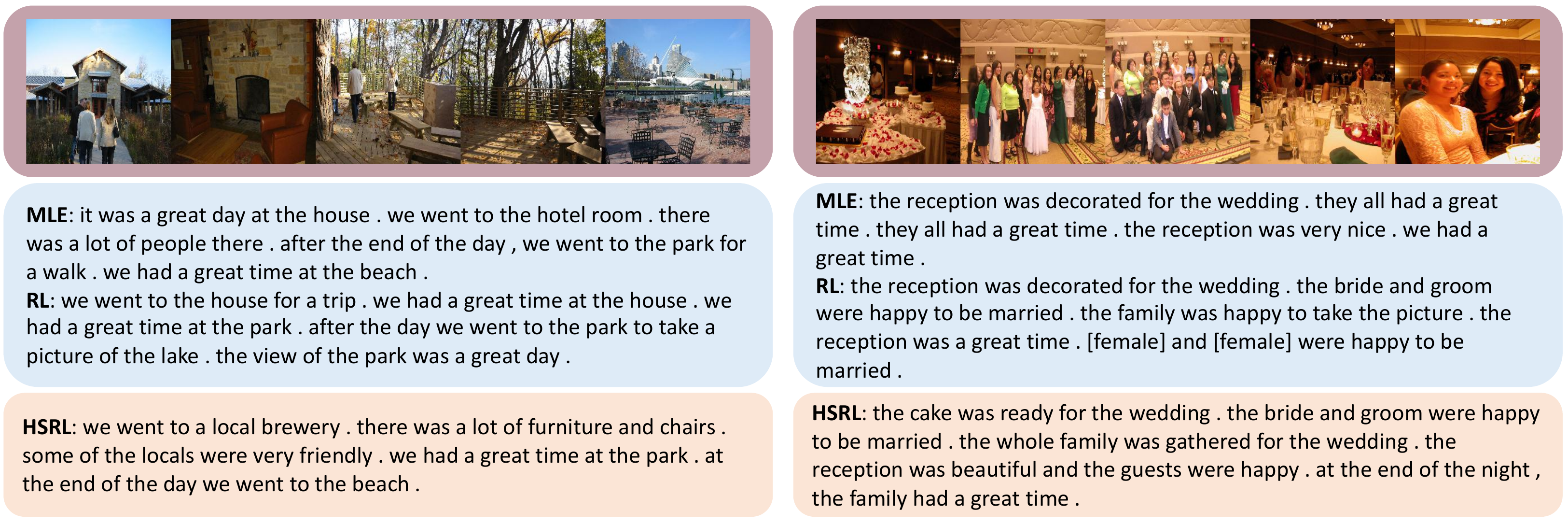} \\
		\includegraphics[width=0.92\textwidth]{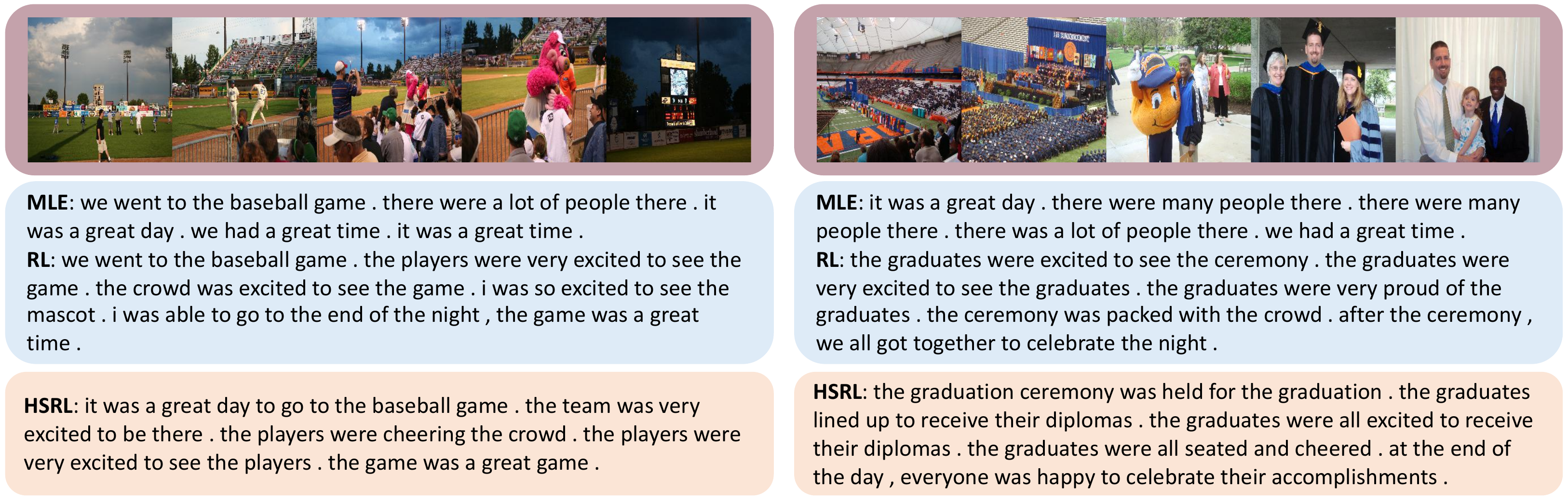}
		\captionof{figure}{Comparison of additional selected stories generated from storytelling models using MLE, RL and our HSRL models given image sequences. 
			\small }
		\label{fig:example_1}
	\end{center}
}]

\clearpage
\twocolumn[{%
	\centering
	\begin{center}
		\centering
		\includegraphics[width=0.92\textwidth]{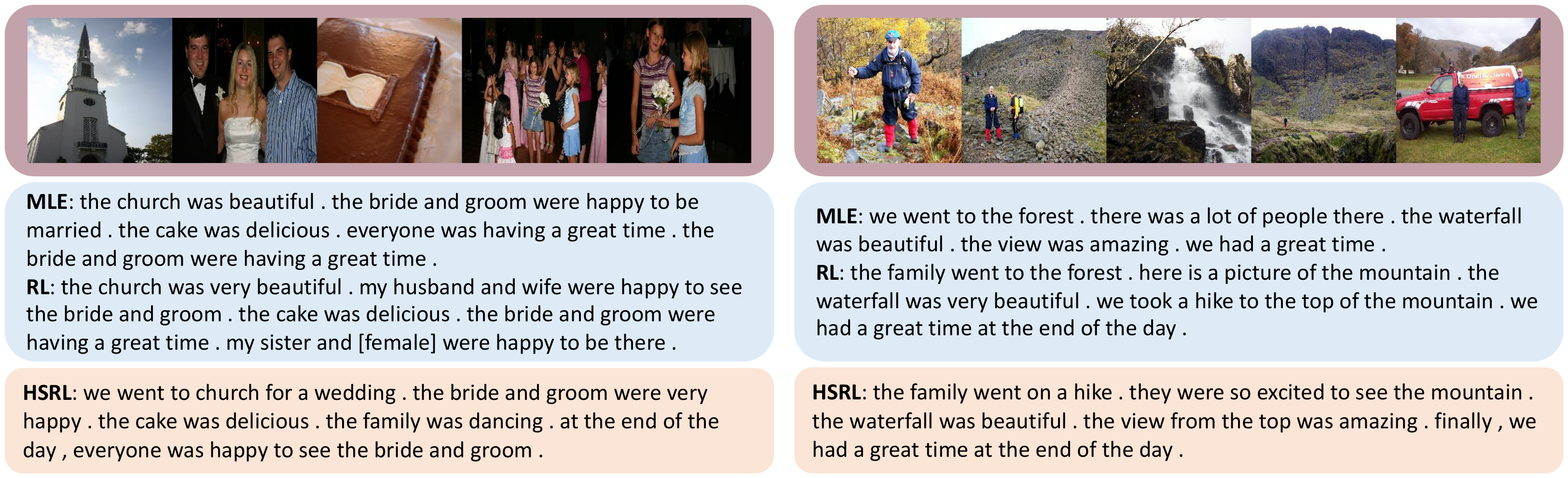} \\
		\includegraphics[width=0.92\textwidth]{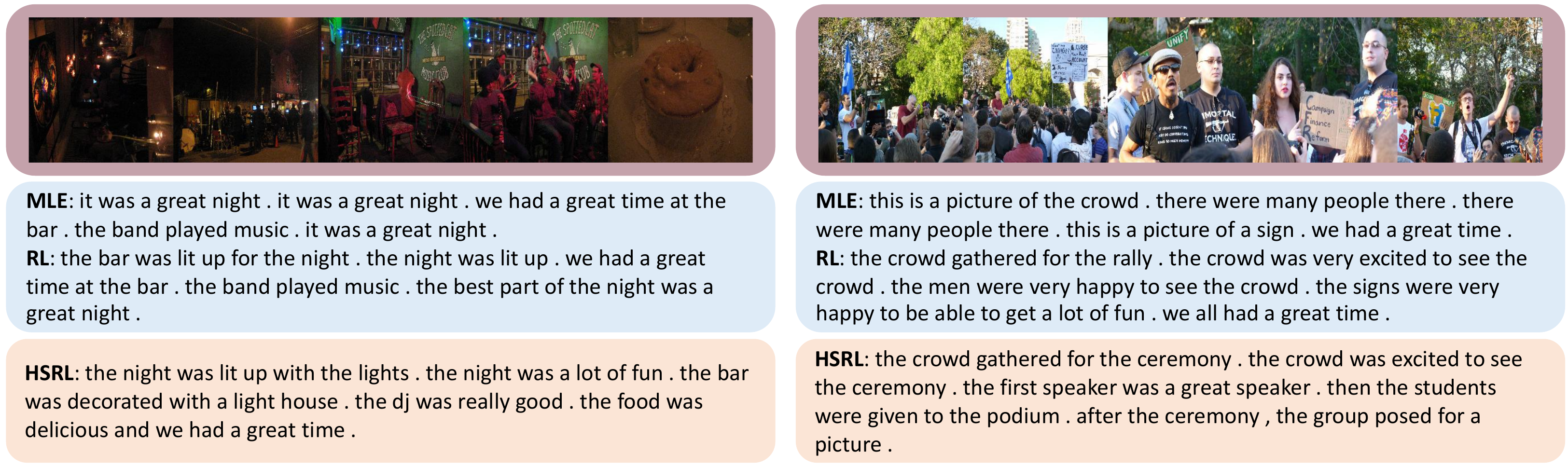} \\
		\includegraphics[width=0.92\textwidth]{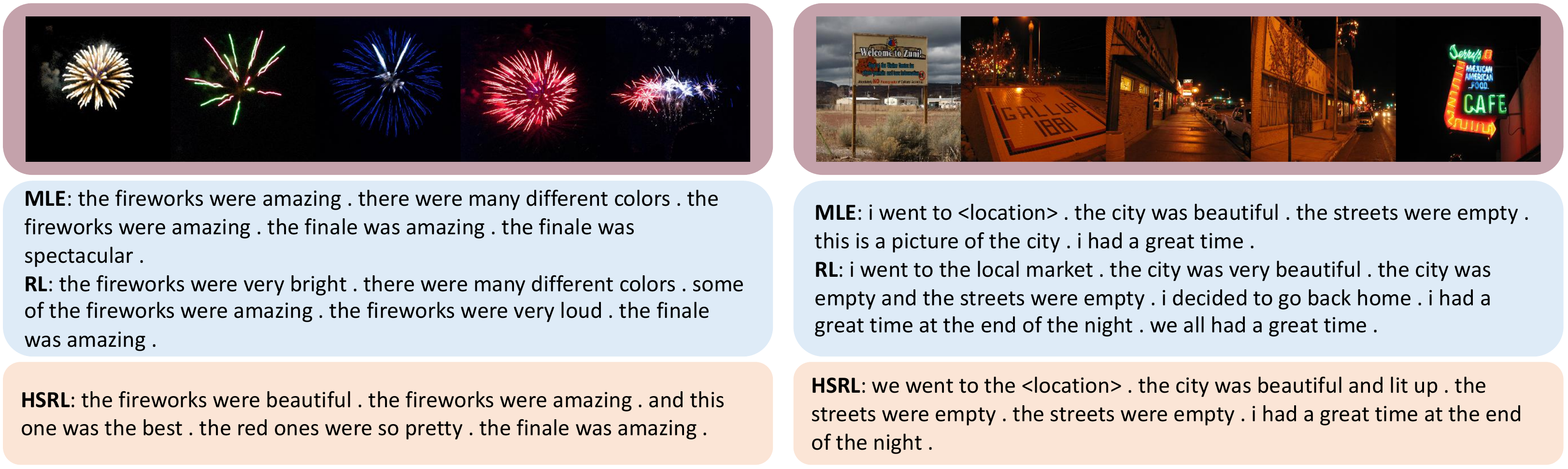}
		\captionof{figure}{Comparison of additional selected stories generated from storytelling models of MLE, RL and our HSRL models given image sequences. 
			\small }
		\label{fig:example_2}
	\end{center}
}]

\paragraph{Impact of $\gamma_1$ and $\gamma_2$} We tried different combinations of $\gamma_1$ and $\gamma_2$ on  the held-out validation set. We found that $\gamma_1=0.7$ best balances the worker and the manager. This is because the overall performance of HSRL depends more on the sentences generated by the worker but less on the topics generated by the manager. Hence, the worker should be more emphasized.  In addition, we found that $\gamma_2=0.9$ optimally balances the RL loss and the MLE loss. CIDEr-D, as an RL reward, is more relevant to the quality of the generated story, and is more correlated to the evaluation metrics indicating that more weight should be placed on the RL loss. 

\begin{figure}[h]
	\centering
	\includegraphics[width=0.35\textwidth]{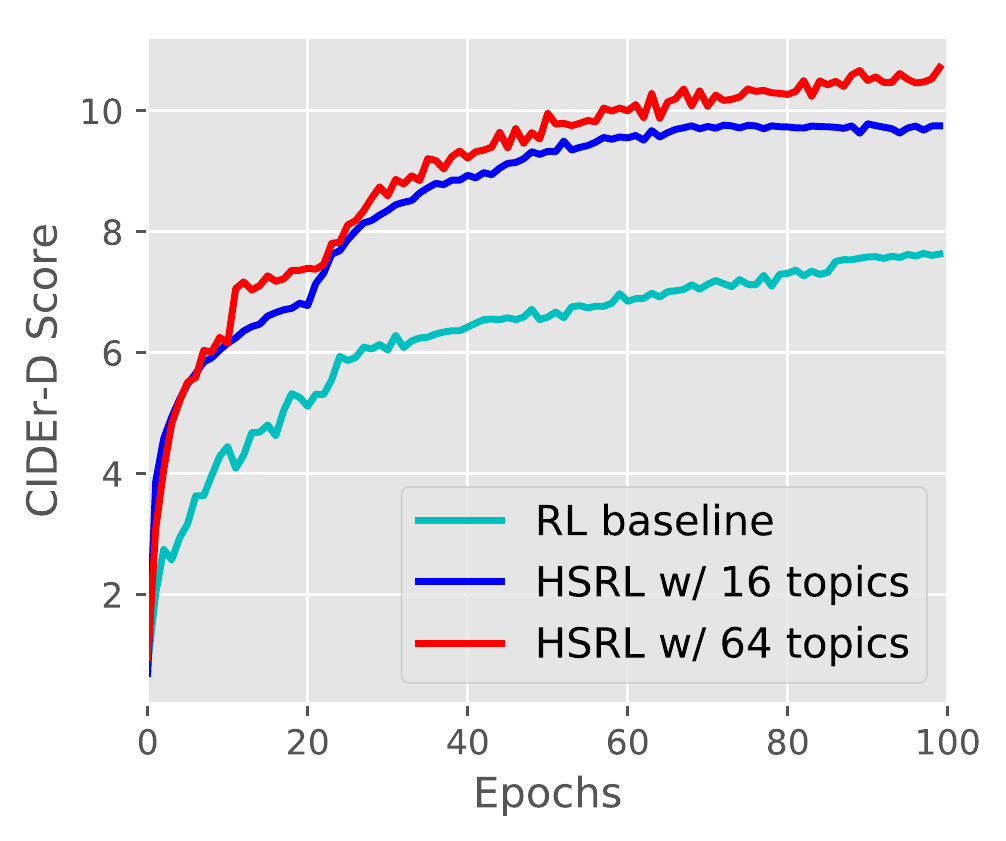}
	\caption{CIDEr-D vs. epoch for an RL baseline and our proposed HSRL approach. \small
	}
	\label{fig:CIDEr}
\end{figure}

\paragraph{CIDEr-D performance effect} 
In Fig.~\ref{fig:CIDEr} the changes in CIDEr-D scores over epochs are shown. We compare the model trained with RL baseline, and our HSRL with 16 and 64 topics. It clearly demonstrates that our HSRL converges with much higher CIDEr result than the RL-baseline LSTM. 
This is explained by the fact that HSRL provides guidance to the workers, while flat RL does not generate goals to see the overall picture of the story.

\end{document}